\documentclass[11pt,a4paper]{article}
\pdfoutput=1 
\usepackage[hyperref]{emnlp2020}
\usepackage{times}
\usepackage{latexsym}
\usepackage{verbatimbox}

\usepackage{graphicx}
\usepackage{subcaption}
\usepackage{multirow}

\usepackage{microtype}

\aclfinalcopy 
\def\aclpaperid{2207} 

\title{TurnGPT: a Transformer-based Language Model for Predicting Turn-taking in Spoken Dialog}

\author{Erik Ekstedt \\
  KTH Speech, Music and Hearing \\
  Stockholm, Sweden \\
  \texttt{erikekst@kth.se} \\\And
  Gabriel Skantze\\
  KTH Speech, Music and Hearing \\
  Stockholm, Sweden \\
  \texttt{skantze@kth.se} \\}

\date{}

\begin{document}
\maketitle
\begin{abstract}

Syntactic and pragmatic completeness is known to be important for turn-taking prediction, but so far machine learning models of turn-taking have used such linguistic information in a limited way. In this paper, we introduce TurnGPT, a transformer-based language model for predicting turn-shifts in spoken dialog. The model has been trained and evaluated on a variety of written and spoken dialog datasets. We show that the model outperforms two baselines  used in prior work. We also report on an ablation study, as well as attention and gradient analyses, which show that the model is able to utilize the dialog context and pragmatic completeness for turn-taking prediction. Finally, we explore the model's potential in not only detecting, but also projecting, turn-completions. 

\end{abstract}

\section{Introduction}

The taking of turns is one of the most fundamental aspects of dialog. Since it is difficult to speak and listen at the same time, the participants need to coordinate who is currently speaking and when the next speaker can start. Traditionally, spoken dialog systems have rested on a very simplistic model of turn-taking, where a certain amount of silence (e.g. 700ms) is used as an indicator that the turn is complete. This often results in interruptions or sluggish responses, depending on where the threshold is set. In human-human interaction, it is clear that much more sophisticated mechanisms are used, where the speakers rely on turn-taking cues (involving prosody and linguistic cues, as well as gaze and gestures) to detect, and even project, turn completions \citep{sacks100247,gravano101679,Levinson2015}. 

More sophisticated models of turn-taking, based on machine learning, have been proposed \citep{Meena2014,Johansson2015,skantze2017,Masumura2019}. Typically, these models rely on the various multi-modal features that have been found to facilitate the coordination of turn-taking. Since dialog is primarily driven by the exchange of meaningful contributions, where each contribution often constitutes some dialog act, linguistic information should intuitively play a major role in turn-taking. However, so far, the representations of linguistic features have been fairly simplistic, and some models rely solely on prosody \citep{Ward2018,Lala2019}. One explanation for this is that the complex semantic and pragmatic functions that the "linguistic cues" should reflect, and which can be expected to regulate turn-taking, are non-trivial for machine learning models to capture, especially since they often depend on the preceding dialog context. 

In this paper, we introduce TurnGPT, a transformer-based language model for turn-taking prediction. Based on Open AI's GPT-2 \citep{radford2019language}, and fine-tuned on various dialog datasets, it predicts possible turn-completion points in dialog, based on linguistic features (words) alone. Transformer-based language models have been shown to perform well on several NLP tasks \citep{radford2019language}. Recent developments in chatbots have also shown that they can produce meaningful utterances in dialog, and thus seem to have a fairly strong representation of the dialog context \citep{transfertransfo}. Through ablation studies and model inspection, we analyse how important the linguistic context is for turn-taking prediction. We evaluate the model using both written and spoken dialog datasets. However, as this paper is focused solely on modelling the linguistic aspect of turn-taking, we do not investigate the contribution of other important features, such as prosody, and leave the combination of such cues with our model for future work. Thus, our  baselines are the linguistic parts of turn-taking models proposed in previous work. 

\section{Background}

One of the most influential early accounts of the organization of turn-taking is the one proposed by \citet{sacks100247}. Their model is based on the observation that since the dialog is not known in advance, it has to be coordinated in a flexible manner as it evolves. Overwhelmingly, one speaker talks at a time; occurrences of more than one speaker at a time are common, but brief. Transitions (from one turn to the next) with very little gap and no overlap are common. Based on these observations, they propose that turns can be constructed from "Turn-constructional units" (TCU). After each such unit, there is a "Transition-relevant place" (TRP), where a turn-shift can (but does not have to) occur, depending on whether the current speaker actively selects the next speaker, or if some other speaker self-selects. 

Several studies have investigated the cues that could be used by the listener to distinguish TRPs ("turn-yielding cues") from non-TRPs ("turn-holding cues") \citep{duncan100200,gravano101679}. For example, in a face-to-face setting, speakers tend to not look at the listener during an utterance, but then shift the gaze towards the addressee when yielding the turn \citep{kendon100127}. Several studies have also investigated prosodic cues for turn-taking, including intonation, duration, loudness and voice quality \citep{Ward2019}. 

From a linguistic perspective, the notion of "completeness" is important, as a complete linguistic unit (such as a sentence) is more likely to be turn-yielding than an incomplete sentence or phrase. \citet{ford100971} analysed linguistic units for turn-taking and proposed two levels of units: syntactic and pragmatic. Syntactic completion, in this context, does not have to be a complete sentence. Neither is a syntactic phrase (like a nominal phrase) necessarily syntactically complete. They define an utterance to be syntactically complete if "in its discourse context, it could be interpreted as a complete clause, that is, with an overt or directly recoverable predicate" (p. 143). This includes "elliptical clauses, answers to questions, and backchannel responses". The syntactic completion is judged incrementally as the utterance unfolds. Figure \ref{fig:trp_example} shows a (made-up) example which illustrates this notion.
\begin{figure}[h]
\begin{verbnobox}[\small]
A: yesterday we met / in the park / 
B: okay / when / will you meet / again /
A: tomorrow / 
\end{verbnobox}
\caption{Example of syntactic completeness (marked by /).}%
\label{fig:trp_example}
\end{figure}
As can be seen, in this account, the turn-initial adverb of time "yesterday" is not syntactically complete (as there is not yet any "overt or directly recoverable predicate"), whereas "tomorrow" is, which illustrates the dependence on the dialog context. As pointed out by \citet{ford100971}, while syntactic completion might be necessary for a TRP, it is not sufficient. Thus, they also introduce the notion of pragmatic completeness, which is defined as "a complete conversational action within its specific sequential context" (p. 150), and corresponds to TRPs. This definition is not very precise, and is likely to depend on a fair amount of common sense. In the example above, while "when will you meet" is syntactically complete, the question is unlikely to end there, given the preceding context, and is therefore not pragmatically complete. 

In their analysis, \citet{ford100971} also argue that the final intonation contour plays an important role in signalling pragmatic completion, where these may be ambiguous. This has also been verified in controlled experiments \citep{Bogels2015}. However, as pointed out by several researchers \citep{Levinson2015,Ward2019}, turn-final prosody cannot (by itself) explain the majority of split-second turn-shifts (around 200ms) that are typically found in data, as the listener would not have time to react, prepare and execute a response. The response time would then be around 600-1500ms \citep{Levinson2015}. Thus, the listener is likely to prepare the response ahead of time and project the turn-completion. For this, they most likely depend on units which are more feasible to project, such as syntactic and pragmatic units. 

Even though syntactic and pragmatic completeness are intuitively important for turn-taking, it is not clear how they should be modelled. So far, most prediction models of turn-shifts have used a very simplistic account of syntactic completion, such as the final part-of-speech tags \citep{gravano101679,Meena2014,Johansson2015}. More recent models of turn-taking have used LSTMs to encode linguistic information, such as part-of-speech \citep{skantze2017}, words \citep{Roddy2018} or senones \citep{Masumura2019}. Although several of these studies have found that linguistic information contribute to the performance (compared to only using prosody), the performance gain is not as big as what could be expected. This calls for the exploration of more powerful linguistic models for  turn-taking.  

\section{Approach}

A problem when modelling TRPs is that they are not overtly present in the data, only actual turn-shifts are. One approach could be to manually annotate TRPs (cf. \citealt{Meena2014,Lala2019}), but this is of course very labour intensive. One could also question the binary notion of TRPs --- a continuous (or probabilistic) notion seems to be more plausible, where transition-relevance varies between highly inappropriate to highly appropriate \citep{Johansson2015}. In this view, a strong TRP should be statistically associated with more turn-shifts. Thus, a probabilistic notion of TRPs should be possible to infer from actual turn-shifts in data, just like a language model (the probability of a word in context) can be inferred from actual language use.  

Given this notion, we include turn-shifts as specific tokens in the vocabulary of a language model and learn their distribution, along with the other tokens, over conversational data in a language model setup. We focus on dialog data and include two separate turn-shift tokens for each of the speakers, which are inserted at the beginning of each speaker turn. A dialog is then a sequence of turns separated by these turn-shift tokens. After training, the probabilities associated to the turn-shift tokens can be viewed as the probability of a TRP. Note, however, that the model not only predicts turn-shifts, but makes predictions over all tokens in the vocabulary, thus retaining its function as a language model.  

The problem of organizing turn-taking primarily concerns spoken language, where response time and fluency has a big impact on the quality of the interaction. However, the process of recording and transcribing spoken dialog is expensive and time consuming. There are also privacy issues regarding recorded speech, which makes audio data less accessible than their written counterpart. Since our focus in this paper is on linguistic aspects of turn-taking, we investigate the use of both written and spoken dialog data. Although the language use is different for spoken vs. written language, we believe that pragmatic TRPs exist and overlap (to some extent) for both types. A clear difference, however, is that spoken language lack punctuation and capitalization, which are not typically available for spoken dialog systems (unless inferred by a transcriber or ASR). Our goal is to learn the distributions over TRPs using linguistic data, without the need to rely on punctuation or capitalization. 
\begin{table*}[]
    \centering
    \begin{tabular}{llllll}
                                                                  &                                  & \#Dialogs                  & \#Turns                     & \#Words/Turn              & \#Unique Words             \\ \hline
        \multicolumn{1}{|l|}{\multirow{4}{*}{Assistant}} & \multicolumn{1}{l|}{Taskmaster}  & \multicolumn{1}{l|}{30.4K} & \multicolumn{1}{l|}{542K}   & \multicolumn{1}{l|}{9.2}  & \multicolumn{1}{l|}{43.7K} \\ \cline{2-6} 
        \multicolumn{1}{|l|}{}                                        & \multicolumn{1}{l|}{MultiWoZ}    & \multicolumn{1}{l|}{10.4K} & \multicolumn{1}{l|}{1.3M}   & \multicolumn{1}{l|}{13.5} & \multicolumn{1}{l|}{18.8K} \\ \cline{2-6} 
        \multicolumn{1}{|l|}{}                                        & \multicolumn{1}{l|}{MetaLWoZ}    & \multicolumn{1}{l|}{37.9K} & \multicolumn{1}{l|}{432K}   & \multicolumn{1}{l|}{7.3}  & \multicolumn{1}{l|}{37.2K} \\ \cline{2-6} 
        \multicolumn{1}{|l|}{}                                        & \multicolumn{1}{l|}{CCPE}        & \multicolumn{1}{l|}{500}   & \multicolumn{1}{l|}{10.1K}  & \multicolumn{1}{l|}{14.4} & \multicolumn{1}{l|}{5K}    \\ \hline
        \multicolumn{1}{|l|}{\multirow{2}{*}{Written Social}}            & \multicolumn{1}{l|}{Persona}     & \multicolumn{1}{l|}{10.9K} & \multicolumn{1}{l|}{162K}   & \multicolumn{1}{l|}{10.1} & \multicolumn{1}{l|}{20.3K} \\ \cline{2-6} 
        \multicolumn{1}{|l|}{}                                        & \multicolumn{1}{l|}{DailyDialog} & \multicolumn{1}{l|}{13.1K} & \multicolumn{1}{l|}{116.4K} & \multicolumn{1}{l|}{10.1} & \multicolumn{1}{l|}{22.2K} \\ \hline
        \multicolumn{1}{|l|}{\multirow{2}{*}{Spontaneous Spoken}}             & \multicolumn{1}{l|}{Maptask}     & \multicolumn{1}{l|}{128}   & \multicolumn{1}{l|}{11.4K}  & \multicolumn{1}{l|}{12.8} & \multicolumn{1}{l|}{2.2K}  \\ \cline{2-6} 
        \multicolumn{1}{|l|}{}                                        & \multicolumn{1}{l|}{Switchboard} & \multicolumn{1}{l|}{2.4K}  & \multicolumn{1}{l|}{106.6K} & \multicolumn{1}{l|}{28.1} & \multicolumn{1}{l|}{27K}   \\ \hline
    \end{tabular}
    \caption{Dataset statistics.}
    \label{tab:datasets}
\end{table*}

\section{Model}%

We use a transformer-based \citep{att_all_need}, uni-directional language model: the GPT-2 \citep{radford2019language} from OpenAI. Transformer models have made a huge impact on NLP research over the past years and was chosen because of their strong performance on language generation. 

Our model can be seen as a modified version of the TransferTransfo \citep{transfertransfo} model, which performed well in the ConvAI2\footnote{http://convai.io/} challenge. In their work, they fine-tuned a GPT \citep{radford2018} model on a particular dialog task with the addition of three tokens, one task-specific and one for each speaker. Transformer-based language models commonly use at least two types of embeddings, a word and a positional embedding. The word embedding encodes the relevant words and the positional encodes their order. TransferTransfo used an additional dialog state embedding consisting of the task specific token and a speaker token for each location, corresponding to the relevant speaker. Training was done using cross-entropy loss and a next-sentence prediction loss. In our work, we omit the task-specific token and the next sentence prediction loss. 

TurnGPT is a GPT2-based transformer using three kinds of embeddings: word, position and speaker id. The speaker tokens are included in the language modelling task and the TRP probability predictions are defined as the maximum assigned output probability over the speaker tokens. Please refer to the code\footnote{https://github.com/ErikEkstedt/TurnGPT} for further details.

We finetune two different pre-trained models, namely GPT-2 \citep{radford2019language} trained on WebText, and DialoGPT \citep{dialogpt} by Microsoft, which is based on GPT-2 but "trained on 147M conversation-like exchanges extracted from Reddit comments". We used the pretrained models available from the transformers \citep{transformers} library using PyTorch \citep{pytorch}. For our experiments, we only used the smallest models (the GPT-2-base and the DialoGPT-small), both with 12 layers, 12 heads and 768 hidden units. 

We compare TurnGPT against two baselines which correspond to linguistic models which have been used in previous research (as reviewed above). First, we train a simple statistical model on part-of-speech (\textbf{POS}) bigrams. For each pair of consecutive POS tags, we get an associated probability of a speaker shift. Second, we train an \textbf{LSTM} model \citep{lstm} with up to three layers with a hidden size of 768. The LSTM baseline is trained directly as a binary turn-shift classifier, given the preceding sequence of words.

\section{Data}
We collect eight dialog datasets with varying characteristics, which we have grouped into three major categories. The first, and largest, group (called \textbf{Assistant}) are task-oriented dialog system corpora, which represent dialog between a user and an automated assistant (where the user typically queries the assistant for information). These datasets were primarily collected through Wizard-of-Oz (WoZ) and self-written dialog (i.e., where one person is writing an imagined dialog), through a crowdsourcing platform: 
\begin{itemize}
    \item The \textbf{Taskmaster} \citep{taskmaster1} dataset (self-written and WoZ using a TTS). 
    \item \textbf{MetaLWOZ}, the dataset for DSTC-8 Track 2 “Fast Domain Adaptation”
        \citep{dstc-8-metalwoz} (WoZ). 
    \item The \textbf{Multiwoz} 2.1 \citep{multiwoz2.1} is an update to the Multiwoz
         \citep{multiwoz} dataset (written WoZ).
    \item The Coached Conversational Preference Elicitation \citep{coached},
        \textbf{CCPE}, dataset (WoZ using a TTS). This
        dataset differs from the previous in that the system tries to extract
        information from the user, as opposed to the other way around.
\end{itemize}        
The second group of datasets (called \textbf{Written Social}) contains human-human written dialogs that are more open and social in nature: 
\begin{itemize}
    \item The \textbf{Persona} dataset \citep{persona} consists of dialogs where two crowdworkers are given the task of trying to get to 
        know each other, based on a given set  of persona attributes (e.g. "I like to ski. I am vegetarian"). 
    \item The \textbf{DailyDialog} dataset \citep{dailydialog} contains dialogs
        extracted from web pages for English learners. The dataset includes a
        variety of topics (relationships, tourism, work, politics, etc). The dataset was intended to resemble dialogs human would have in their "daily life".
\end{itemize} 
The third type of collected data is that of \textbf{Spontaneous Spoken} dialog between two humans:
\begin{itemize}
    \item \textbf{Maptask} \citep{maptask} is a task-oriented dataset
        where a "guide" explains a defined route on a map to a "follower" which
        tries to draw that path on their map.  
    \item \textbf{Switchboard} \citep{swb} contains more open-ended telephone dialogs, constrained only by a given topic (e.g. recycling).
\end{itemize}
Table \ref{tab:datasets} shows the basic statistics over each of the eight datasets. All datasets were also combined to create a \textbf{Full} dataset. Each dataset was split into training, validation and test sets, using predefined splits if available, or else a random split of (90/5/5). 

\subsection{Data Extraction}%

The dialogs were extracted from the different corpora. For each turn, a speaker token was inserted at the start (\texttt{speaker1} or \texttt{speaker2}). Punctuations (\texttt{,.:;!?}) were removed, and all characters were made lower case. The words were encoded by a bytepair encoding (BPE) vocabulary \citep{sennrich} used in the respective pretrained models. This method splits words into subwords, and the final vocabulary consists of 50,261 tokens.

For all datasets, except Maptask and Switchboard, the turns were explicitly given by the structure of the data. Since the Spontaneous Spoken datasets contain a fair amount of overlap and have no clearly defined turns, a custom turn extraction policy was implemented: First, backchannels were defined by a set of candidates (e.g "mm", "mhm", etc) and removed from the dialog if they were spoken in isolation, separated by more than a second from other utterances made by the same speaker. Second, IPUs (Inter-pausal units) were defined as utterances separated by less than 500ms. IPUs of one speaker, spoken completely inside an IPU made by the other speaker, were omitted. Third, a sequence of turns was created by merging all consecutive IPUs from one speaker, separated by mutual silence, into a single turn. The turns were ordered by time, ignoring any overlap between them. These turns were then treated the same way as for the rest of the datasets.

For the POS baseline we used the NLTK \citep{nltk} library to extract POS tags from the extracted dialogs. 

\section{Experiment}%
\subsection{Training}%
We trained the models on both the Assistant and Full training sets using the cross-entropy loss. The models with the lowest validation loss were then used for testing. The TurnGPT models used the AdamW optimizer and the default hyperparameters of the transformer library.

The LSTM baseline used the same tokens provided by the GPT-2 model but trained on the binary prediction of the next token being a turn-shift or not, using a sigmoidal output activation on the mean squared error loss. The LSTM model utilized the AdamW optimizer included in PyTorch, with a weight decay of 0.01, dropout of 0.1, and a learning rate of 6.25e-5. We used up to three hidden layers for the LSTM and chose the one that performed best on the validation sets, which was the 2-layer LSTM.

\subsection{Evaluation}%

The best performing models on the validation sets were used to evaluate the performance on the test sets. Each model have associated probabilities related to turn-shifts. For the transformer-based models, we chose the maximum speaker token output probability at each time step as the probability of a TRP. Since the LSTM baseline was directly trained as a turn-shift classifier, the output could be used directly as a TRP probability. The POS baseline follows the same reasoning. The chosen evaluation critera was the balanced accuracy (bAcc) over true and false turn-shifts. This metric was chosen because of the imbalanced classes (there are many more word tokens than turn-shifts). The bAcc is defined by
\begin{equation}
    bAcc = \frac{TPR + TNR}{2}  \in [0.5, 1],
\end{equation}
where $TPR$ and $TNR$ is the true positive rate (positive recall) and the true negative rate respectively. The lowest value is 0.5, which is achieved by always guessing on one class, and the highest is 1 (100\% accuracy over both classes). 

To use the models as classifiers, a threshold was used to discretize the probabilities into two classes, where a probability over the threshold was regarded as a turn-shift. We used independent thresholds for each model that yielded the highest test score.
The results are shown in Table \ref{tab:results}. As can be seen, the TurnGPT models achieved the best results on all datasets. Both GPT-2 and DialogGPT yielded similar performance. When evaluated on the Spoken and Written datasets, the models also benefit from training on the Full dataset (where these are included). This shows that the language use across these datasets indeed differ, and that it is important to train the models on different types of corpora. Overall, turn-shift predictions on the Spoken and Written datasets are more challenging, which can be explained by their more spontaneous nature. 

\begin{table}[]
\begin{tabular}{ll|l|l|l|}
\cline{3-5}
                                                 &                          & Assistant       & Spoken          & Written         \\ \hline
\multicolumn{1}{|l|}{\multirow{4}{*}{\rotatebox{90}{Assistant}}}& POS                      & 0.696           & 0.659           & 0.733           \\ \cline{2-5} 
\multicolumn{1}{|l|}{}                           & LSTM                     & 0.866           & 0.690           & 0.795           \\ \cline{2-5} 
\multicolumn{1}{|l|}{}                           & \multirow{2}{*}{TurnGPT} & \textbf{0.913}  & \textbf{0.789}  & 0.875           \\ \cline{3-5} 
\multicolumn{1}{|l|}{}                           &                          & 0.912*          & 0.784*          & \textbf{0.877*} \\ \hline
\multicolumn{1}{|l|}{\multirow{4}{*}{\rotatebox{90}{Full}}}      & POS                      & 0.750           & 0.675           & 0.732           \\ \cline{2-5} 
\multicolumn{1}{|l|}{}                           & LSTM                     & 0.869           & 0.748           & 0.83            \\ \cline{2-5} 
\multicolumn{1}{|l|}{}                           & \multirow{2}{*}{TurnGPT} & \textbf{0.913}  & \textbf{0.823}  & 0.905           \\ \cline{3-5} 
\multicolumn{1}{|l|}{}                           &                          & \textbf{0.913*} & \textbf{0.823*} & \textbf{0.906*} \\ \hline
\end{tabular}
\caption{The bAcc on different test sets, with models trained on the Assistant and Full training sets. TurnGPT entries with (*) indicates the DialoGPT version.}%
\label{tab:results}
\end{table}

\begin{figure}[h]
    \centering
    \includegraphics[width=0.95\linewidth]{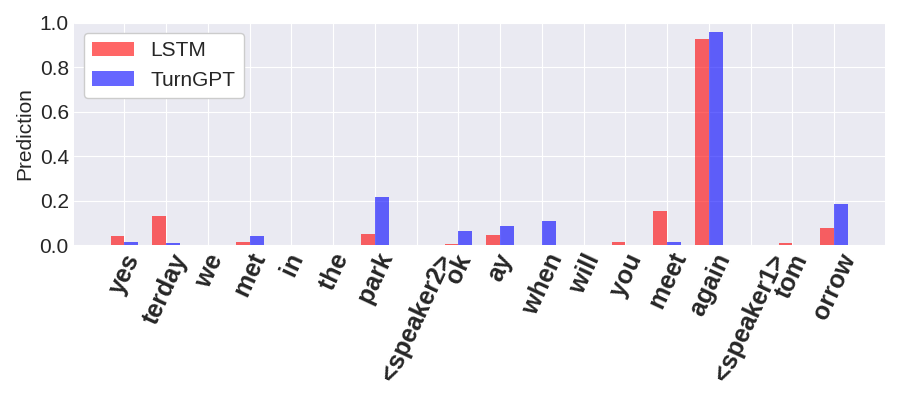}
    \caption{TRP probabilities associated with the constructed sample in Figure \ref{fig:trp_example}.}%
    \label{fig:trp_model_sample}
\end{figure}

A sample visualization over the TRP probabilities for the example in Figure \ref{fig:trp_example}, as yielded by the LSTM and TurnGPT models, is shown in Figure \ref{fig:trp_model_sample}. First, this figure shows how a more probabilistic notion of TRPs is intuitively more compelling than a binary notion. Second, the example clearly illustrates some of the benefits of the TurnGPT model over the LSTM model. The LSTM model gives a fairly high probability of a TRP after "yesterday", and somewhat high after "tomorrow". Without considering the previous context, these should indeed be fairly equivalent. The TurnGPT model, on the other hand, correctly separates these two instances, presumably because it has a better model of the context.  Similarly, the LSTM model (but not TurnGPT) assigns a fairly high probably for a TRP after "when will you meet", indicating that it is sensitive to syntactic, but perhaps not pragmatic, completeness, in the sense of \citet{ford100971}.

\subsection{Context Ablation}
\begin{figure}[h]
    \centering
    \includegraphics[width=0.95\linewidth]{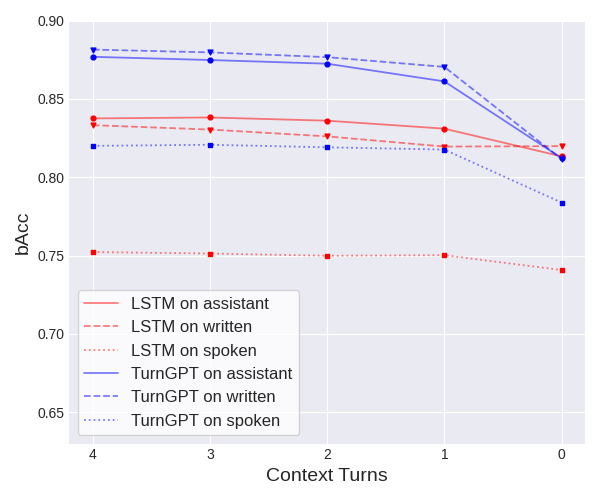}
    \caption{The bAcc score for the TurnGPT model and the LSTM baseline trained on the full dataset.}
    \label{fig:context_ablation}
\end{figure}

In order to bring further insight into the importance of context, we perform an ablation study, varying the amount of context available to the model. For this, we only use turns that have a minimum of 4 preceding turns. For context 0, only the current turn is given as input, but for context 4, the current turn and the 4 preceding turns are used as input. The evaluation is done over all suitable turns. The results are shown in Figure \ref{fig:context_ablation}. 

For TurnGPT, the performance increases with the amount of context. The biggest drop in performance happens when going from some context to no context. We note that the LSTM classifier shows a similar behaviour, but to a less extent, and actually improves the performance slightly when going from a single context turn to zero on the Written dataset. 

\begin{figure}[h]
    \centering
    \includegraphics[width=0.99\linewidth]{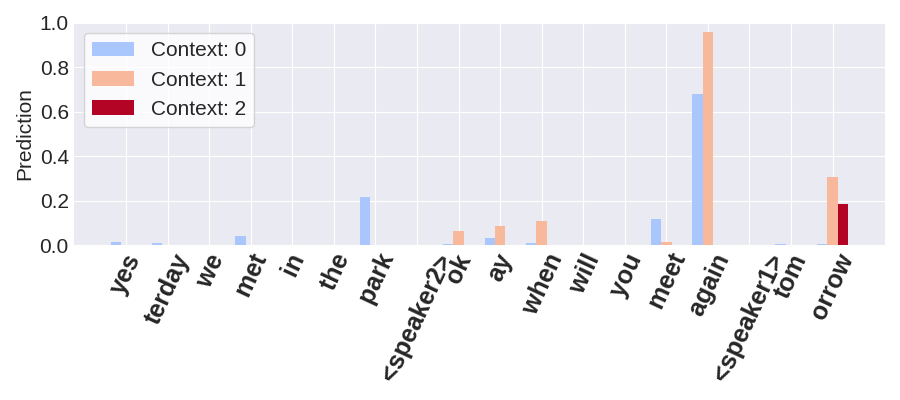}
    \caption{TurnGPT predictions for varying context over the constructed sample in Figure \ref{fig:trp_example}. The blue bars are only given the current turn as input. The orange bars further includes the previous context turn and the red includes the two previous turns (which is only relevant for the last turn).
    }%
    \label{fig:trp_context_sample}
\end{figure}

To visualize how the TurnGPT model might change its prediction depending on the available context, we include a visualization over the constructed sample in Figure \ref{fig:trp_context_sample}. After the last word "tomorrow", we note how having no context vs. some context changes the prediction for a turn-shift considerably. In other words, the model has learned that a turn-initial "tomorrow", by itself, is very unlikely to be the end of a turn. However, interpreted in the context of the preceding question, the probability is much higher.

\subsection{Model Inspection}

We further investigate the contextual impact by looking at the attention mechanism inherent in any transformer-based model. Inspired by the work of \citet{WhatDoesBertLookAt}, we extract the attention over all true turn-shift tokens where the model assigned a turn-shift probability over 0.2. We added together the attention contribution over each of the 5 most recent turns (the current turn and 4 context turns). 

\begin{figure}[h]
    \centering
    \includegraphics[width=0.99\linewidth]{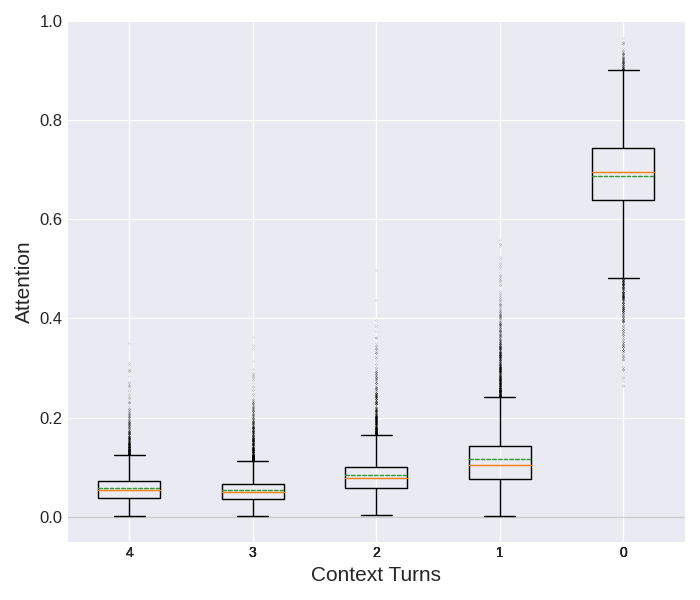}
    \caption{The normalized distributions over turn attention for the last five turns, including the current. 
    }%
    \label{fig:aggregate_attention}
\end{figure}

The output token part of the model may attend to all previous tokens (including itself). Each turn has varying amounts of preceding tokens, and to better understand the attention over the most recent context, we normalize over the 5 most recent turns. In other words, the attention contributions for the last 5 turns will sum up to 1 (for any individual sample). The distributions over turn attention is shown in Figure \ref{fig:aggregate_attention}. The current turn contains, on average, around 70\% of the contextual attention, which is reasonable given that most information regarding turn-shifts are expected to be in the current turn. The remaining 30\% still constitute a substantial contribution, which further strengthen the conclusion that dialog context is important.  

In addition to the attention we investigate the importance the model puts on the last 5 turns by calculating the Integrated Gradient (IG) \citep{ig}. The integrated gradient technique is useful for investigating the effect the input has on any particular output. In this case that can be interpreted as how much any word contributes towards a turn-shift prediction. 

As further described in \citet{ig}, this method requires a definition of a baseline.
We tried the recommended zero word vector as a baseline, but found that the \texttt{unk} (unknown) token worked better. The speaker tokens were considered fixed and were kept intact in the IG calculations.

\begin{figure}
    \centering
    \includegraphics[width=0.99\linewidth]{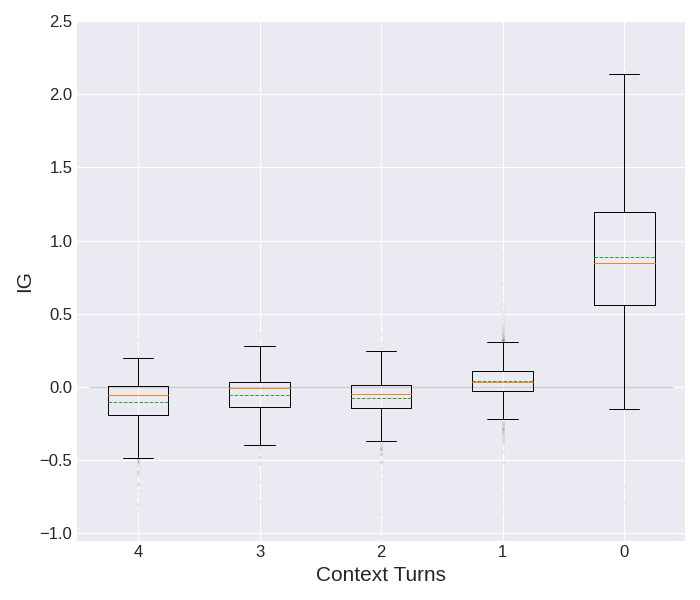}
    \caption{The distributions over the integrated gradient turn sum of the last five turns including the current. The gradient was calculated with respect to the last token in the current turn.}%
    \label{fig:gradient_context_aggregate}
\end{figure}

We are interested in the model's behaviour when it predicts a turn-shift to be likely. We chose only targets at true turn-shift locations with a predicted turn-shift probability over 0.2, the same value used in the attention analysis. The IG contribution values were averaged over each of the 5 most recent turns. Because this approach requires much computation, we randomly chose 500 dialogs from the full test set and calculated 2 targets in each dialog for a total of 1000 integrated gradients. The results are shown in Figure \ref{fig:gradient_context_aggregate}. 

The integrated gradient shows both positive and negative contributions. The first turn is mostly positive and indicates that the immediate context contributes, on average, positively towards predicting a turn-shift. For example given that the last words form a question, each of the "question" words arguably contribute positively towards a turn-shift. 

However, the preceding turns contribute more negatively, thus decreasing the likelihood of a turn shift at the target. One potential explanation for this is that the context provides evidence that a syntactic completion is not a pragmatic completion. However, this hypothesis needs to be investigated further in future work. 

\subsection{Future Prediction}

An interesting aspect of learning the distributions of turn-shifts through a language model setup is the ability to generate text and inspect possible futures. This is done by sampling from the output distribution of the model in an autoregressive manner, and then count the number of tokens until a generated speaker-token. In a dialog system, this would allow the model to estimate the time until a turn completion, and thereby open up for models that can project (and not just detect) turn completions. This would give the system more time to prepare a response and be able to respond with almost no gap, similar to human-human dialog. 

Although we leave this to be further explored in future work, we perform a simple experiment here to evaluate the feasibility of the idea. As an example, we again use the dialog from Figure \ref{fig:trp_example}, and sample over the cumulative output distribution under 0.9, for a maximum length of 50 tokens. 

\begin{figure}[h]
    \centering
    \includegraphics[width=0.99\linewidth]{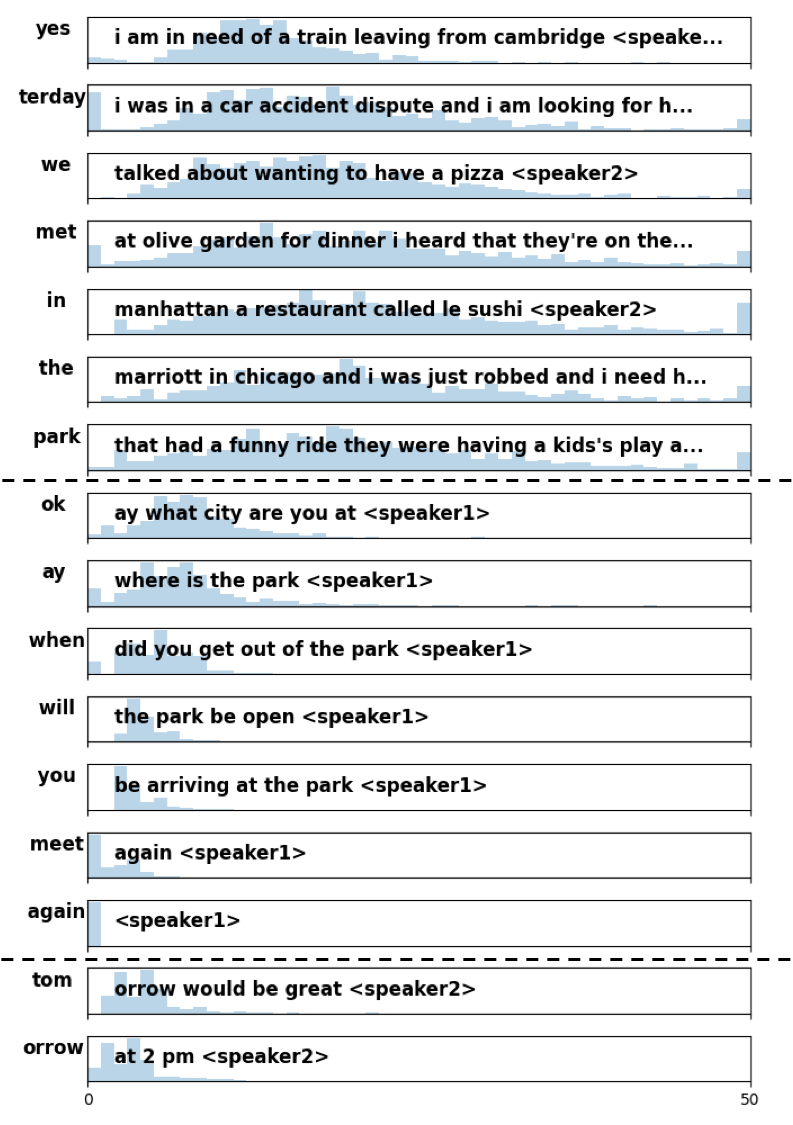}
    \caption{Histograms over predicted turn lengths with a generated sample shown as text. The text may be read from the token on the first y-axis down to any token of interest and then continue reading left to right. Turns are separated by the dashed lines.}%
    \label{fig:future_generation}
\end{figure}

The histograms in Figure \ref{fig:future_generation} show the predicted number of tokens left in the turn, generated over 1000 samples. We note that during the first turn, the model is biased to produce longer sequences, as there is no context that provides any constraints. However, already in the second turn this behaviour changes, and the predictions become much shorter, which further adds to the notion that turn-shift prediction is informed by context. In this specific example, we also note that the predicted turn completions decrease in length and becomes more stable the closer we get to the end of the turn. 

\section{Conclusions and Discussion}

In this paper, we presented a model for turn-shift prediction by formulating the problem as a language modelling task. We introduced TurnGPT, a model which is a finetuned GPT-2 transformer imbued with special turn-shift tokens. The model performed better than baselines used in previous work. Through an ablation study and model inspection, we showed that this is partly thanks to the strong representation of context that prior models lack, i.e., the model's ability to identify pragmatic (and not  just syntactic) completion. We also showcased the model's ability to generate possible futures as a way of predicting upcoming turn-shifts. 

As we are addressing spoken dialog, this work should be seen as an important step towards a more powerful turn-taking model that takes both linguistic information, as well as prosody and other cues (such as gaze and gestures) into account. As argued in the linguistic literature \citep{ford100971,Bogels2015}, prosodic information can be important to further disambiguate pragmatic completion. However, we argue that previous models that have combined linguistic and prosodic cues (cf. \citealt{Meena2014,skantze2017,Roddy2018,Masumura2019}) have used too simplistic models of linguistic turn-constructional units. The integration of prosodic information with a model like TurnGPT is an important topic for future work.   

TurnGPT could also be interesting not just from a dialog system perspective; further model inspection and ablation studies could also be used to identify more exactly how certain words contribute to turn-completion predictions. This can potentially give insights into how humans manage to coordinate their turn-taking in spoken interaction with each other.

\section*{Acknowledgements}
This work is supported by the Swedish research council (VR) project CORDIAL (\#2013-1403) and the SSF project COIN.

\bibliographystyle{acl_natbib}
\bibliography{references}


\end{document}